# EXPERTNet: Exigent Features Preservative Network for Facial Expression Recognition

Monu Verma
monuverma.cv@gmail.com

Jaspreet Kaur Bhui
jassijkboo1@gmail.com

Santosh K. Vipparthi
skvipparthi@mnit.ac.in

Girdhari Singh
gsingh.cse@mnit.ac.in

**Dept. of Computer Science and Engineering, Malaviya National Institute of Technology, Jaipur, India**

## ABSTRACT

Facial expressions have essential cues to infer the humans state of mind, that conveys adequate information to understand individuals' actual feelings. Thus, automatic facial expression recognition is an interesting and crucial task to interpret the humans cognitive state through the machine. In this paper, we proposed an Exigent Features Preservative Network (EXPERTNet), to describe the features of the facial expressions. The EXPERTNet extracts only pertinent features and neglect others by using exigent feature (ExFeat) block, mainly comprises of elective layer. Specifically, elective layer selects the desired edge variation features from the previous layer outcomes, which are generated by applying different sized filters as $1 \times 1$, $3 \times 3$, $5 \times 5$ and $7 \times 7$. Different sized filters aid to elicits both micro and high-level features that enhance the learnability of neurons. ExFeat block preserves the spatial structural information of the facial expression, which allows to discriminate between different classes of facial expressions. Visual representation of the proposed method over different facial expressions shows the learning capability of the neurons of different layers. Experimental and comparative analysis results over four comprehensive datasets: CK+, MMI DISFA and GEMEP-FERA, ensures the better performance of the proposed network as compared to existing networks.

## 1 Introduction

Human emotions indicate their intentions, mental state and feelings, which can be expressed through speech, gestures and facial expressions. Facial expressions have sufficient information itself to understand the psychological state of a person. Each expression forms a unique pattern on face. For human beings it becomes unexacting to identify the expressions due to continuous learning process of brain which starts from their birth. But it is a challenging task to do same for machines. With recent advancement in computer vision and machine learning techniques, it has become possible up to some extent, but still needs some mutative improvements in existing approaches. Paul Ekman [1] presented a thorough study of human emotions. They classified human emotions into six basic categories namely: anger, disgust, fear, happy, sad and surprise. Additionally, Ekman and Friesen [2] developed facial action coding system (FACS) which was further interpreted as emotional facial coding system (EFACS) [3]. EFAC provide a standard way to analyze the facial expressions that achieve extensive results compare to previous approaches. However, there are many variants available like illumination changes, head pose variation, age, ethnicity, etc. which become hurdles to develop a robust facial expression recognition (FER) system.

A general flow of designing an automated FER system takes three steps: face acquisition, developing feature extraction techniques and expression classification. In face acquisition step, facial images are augmented by cropping the facial regions and subtracting unnecessary background noise from an input image, which can create uncertainty during recognition process. After face acquisition salient features are extracted from preprocessed images, which are responsible for generating edge patterns and allows to discriminate between expression classes. Furthermore, Feature extraction techniques can be divided into two categories that are predesigned and learning based techniques. Predesigned techniques utilized handcrafted filters to capture pertinent feature response of facial appearance. These can be further classified into two categories geometric and appearance-based recognition. Geometric based techniques [4, 5] identify position and shape of the Action Units (AUs), then represented them through the feature vectors. But, these techniques fully rely on adequate facial point selection, which are difficult to detect under appearance changes on the face for different expression classes. However, appearance-based approaches are uses handcrafted filters such as Gabor Filters [6], Local Binary Pattern (LBP) [7-8], Local Directional Patterns (LDP) [9], Local Directional Number Pattern (LDN) [10], local Directional Texture Pattern (LDTexP) [11], Local Directional Ternary Pattern (LDTerP) [12] etc., to represents the facial expression features. LBP and its variants have yielded impressive accuracy, but they are sensitive to noise and are unable to handle intra-class variations. Furthermore, Rivera et al. [9] introduced the concept of directional pattern, they encoded feature through directional index value instead of actual intensity value and gain good results.

Nowadays, with rapid development of technologies as machine learning, artificial intelligence, computer vision, hardware designs (Graphical Processing Unit: GPU), learning based techniques got intensive attention. Mainly, Convolutional Neural Networks (CNNs) shows the remarkable improvements in field of image classification. Deep learning approaches learn both the feature extraction and network weight parameters for accurate

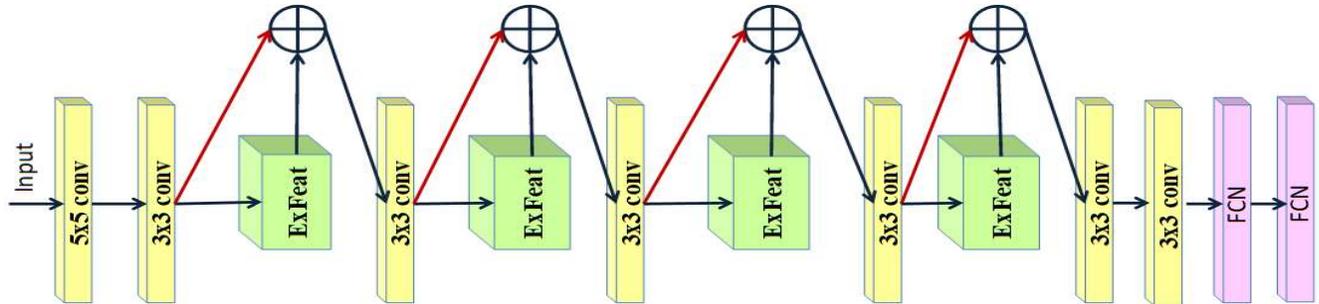
Figure 1: Proposed EXPERTNet Architecture

classification. Many deep CNN models are proposed in literature such as AlexNet [13], VGG-Net [14], GoogleNet [15] ResNet [16] etc.

Earlier, Krizhevsky et al. [13] introduced a network named as AlexNet, which utilizes the concept of drop out layer to reduces the overfitting problem. They also enhance the capability of network by extending dataset through data augmentation. Furthermore, VGG Net [14] have achieved good performance in the Image Net Large Scale Visual Recognition Challenge (ILSVRC) in 2014. After that, Szegedy et al. [15] designed a novel deep network with 22 CNN layers as GoogLeNet, which has incorporated with inception module. ResNet [16], network is the extension of the VGG Net along with residual layer. Residual layer is the core of the ResNet, which combines the outcome of processed layer with the prior layer and enhance the quality to edge pattern responses. These edge pattern responses improve effectiveness of the network by providing adequate features to neurons. Mollahosseini et al. [17] adopted the inception layer [15] concept to train FER system and observed that, it enhances the automatic learnability of the neurons. To improve the robustness of the network, Hassani et al. [18] introduced a spatio-temporal net with conditional random field. This network extracts both appearance and momentary variations features to represent the facial expression images. Furthermore, Jung et al. [19] a fused CNN network named as DTGAN, which used the joint fine tuning to represents the facial appearance features. DTGAN network is a combination of two networks: first network extracts the geometric features by using landmark points and another network represents the appearance features of the facial expressions. Khorrami et al. [20] introduced a zero bias CNN network along with augmentation and drop out layer (Zero bias CNN and AD), to capture the salient features of the expressive facial images. Furthermore, Ding et al. [21] propose a FaceNet2ExpNet, which utilized the transfer learning method. This network comprises two networks as FaceNet and ExpNet. FaceNet is trained over face images and then learned weights are feed to the ExpNet. Further, ExpNet is trained for the expression classification. Kim et al. [22] proposed a deep generative contrastive network (GCNet), which embedded addition encoder and decoder layers in a network to represents the contrastive structure to differentiate the expression classes. Burket et al. [23] introduced a DeXpression network, that consists of two feature extraction blocks. First block holds identical filter size which incorporates the accurate sparseness in the network. Another block has different sized filters to capture both minor and major edge variations from the face images. Later, Zang et al. [24] proposed a deep evolutional spatial-temporal network, which utilized the advantages of two networks: recurrent neural network and multi-signal CNN to analyzed the dynamic evolution and appearance information of the facial expression respectively.

Motivated by former networks, in this paper, we proposed an Exigent Feature Preservative Network to preserve only desired features and neglects the unnecessary ones. The main contribution of the proposed network is summarized as follows:

1. An Exigent Feature Preservative Network (EXPERTNet) with two main blocks: Exigent Feature (ExFeat) block and Additive layer, is proposed to represent spatial structure of the expression appearance.
2. ExFeat block, mainly comprises of elective layer, that extracts the pertinent features from both micro and high-level feature responses generated by different sized filters at convolution layer. Elective layer also improves performance of the network by reducing the learning parameters of the hidden layers.
3. Additive layer integrated salient features of the two layers and holds hybrid response features, which have high-quality features and allows classifier to make an effective decision.

The performance of the proposed network is measured over four standard datasets: CK+, MMI, DISFA and GEMEP-FERA. From the experimental results and comparative study, it is clear that, EXPERTNet achieves impressive results with fair margins as compared to existing state-of-art approaches.

## 2 Proposed Architecture

Inspired by the literature [15, 16], a novel CNN-based model Exigent Features Preservative Network (EXPERTNet) is proposed. EXPERTNet mainly comprises of four parts: convolution layer, additive layer, exfeat block and fully connected layer as shown in Figure 1. The network starts with two consecutive convolution layers (Conv 1 $\rightarrow$ ReLU 1 $\rightarrow$ Conv 2 $\rightarrow$ ReLU 2) with 32 filters of size $5 \times 5$ and $3 \times 3$ respectively.

**Convolution Layer:** Convolution layer imposes learnable filters over input image by applying dot product and extracts response feature map. Each layer contains some specific neurons with learnable weights and bias, which are responsible for capturing pertinent features from the input. Let I (p × q) be an input image and $f_k$ (u × v), $k \in N$, represents the convolved filter with kernel size (u × v). N implies for depth of the filters. The response feature map is computed by Eq. 1.

TABLE I
EXPERTNET DETAILED CONFIGURATION

| Layers | | Filter | Output | # Param |
|---|---|---|---|---|
| Input Image | | - | $128 \times 128 \times 3$ | - |
| Conv 1 | | $5 \times 5$ | $128 \times 128 \times 32$ | 2K |
| Conv 2 | | $3 \times 3$ | $64 \times 64 \times 32$ | 9K |
| ExFeat 1 | Conv 3.1 | $1 \times 1$ | $64 \times 64 \times 32$ | 86K |
| | Conv 3.2 | $3 \times 3$ | | |
| | Conv 3.3 | $5 \times 5$ | | |
| | Conv 3.4 | $7 \times 7$ | | |
| Addition 1 | | - | $64 \times 64 \times 32$ | - |
| Conv 4 | | $3 \times 3$ | $32 \times 32 \times 64$ | 18K |
| ExFeat 2 | Conv 4.1 | $1 \times 1$ | $32 \times 32 \times 64$ | 342K |
| | Conv 4.2 | $3 \times 3$ | | |
| | Conv 4.3 | $5 \times 5$ | | |
| | Conv 4.4 | $7 \times 7$ | | |
| Addition 2 | | - | $32 \times 32 \times 64$ | - |
| Conv 5 | | $3 \times 3$ | $16 \times 16 \times 96$ | 55K |
| ExFeat 3 | Conv 6.1 | $1 \times 1$ | $16 \times 16 \times 96$ | 773K |
| | Conv 6.2 | $3 \times 3$ | | |
| | Conv 6.3 | $5 \times 5$ | | |
| | Conv 6.4 | $7 \times 7$ | | |
| Addition 3 | | - | $16 \times 16 \times 96$ | - |
| Conv 7 | | $3 \times 3$ | $8 \times 8 \times 128$ | 111K |
| ExFeat 4 | Conv 8.1 | $1 \times 1$ | $8 \times 8 \times 128$ | 1M |
| | Conv 8.2 | $3 \times 3$ | | |
| | Conv 8.3 | $5 \times 5$ | | |
| | Conv 8.4 | $7 \times 7$ | | |
| Addition 4 | | | $8 \times 8 \times 128$ | - |
| Conv 9 | | $3 \times 3$ | $4 \times 4 \times 184$ | 212K |
| Conv 10 | | $3 \times 3$ | $2 \times 2 \times 256$ | 424K |
| Fully Connected 1 | | - | $1 \times 1 \times 512$ | 525K |
| Fully Connected 2 | | - | $1 \times 1 \times 1024$ | 525K |

$$R_{Conv} = \sum_{m=1}^{v} \sum_{n=1}^{u} f_k(m,n) \otimes I(p-m, q-n) \quad (1)$$

In EXPERTNet, we utilized the convolution with stride 2 for the downsampling of the input image instead of max pooling layer. Max pooling uses the max operation over the filtered responses that neglects the micro-level edge information, which has key role in expression analysis. Convolution layer performs cross-correlation and secures more salient features. The response of convolution with stride 2 is generated by using Eq. 2-3.

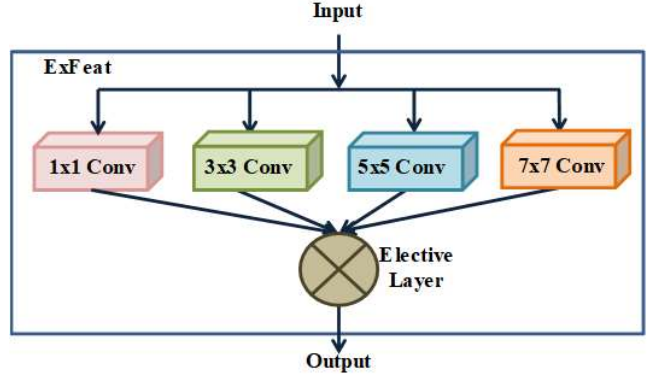

Figure 2: Detailed structure of ExFeat block

$$R_{Conv/2} = \sum_{m=-v/2}^{v/2} \sum_{n=-u/2}^{u/2} f_k(m,n) \otimes I(\alpha - m, \beta - n) \quad (2)$$

Where,

$$\begin{cases} \alpha = 2p - 1 \\ \beta = 2q - 1 \end{cases} \quad (3)$$

**Rectified Linear Unit:** ReLU layer is used to transformed linear input into nonlinear by imposing a monotonic function. ReLU extends the capability of the EXPERTNet by dropping gradients, which are not active in the network. First, Alexnet [13] architecture uses ReLU function in-place of tanh activation, to generates the sparse feature responses. The outcome of ReLU is computed by using max function as:

$$R_\psi = \max(0, I(p \times q)) \quad (4)$$

Furthermore, the outcome feeds to the ExFeat block, that preserves the pertinent features of facial appearance structure from the facial image.

**ExFeat Block:** ExFeat block is created by inspiring from the inception layer [15]. This block comprises four convolutions and one elective layer. Convolution layers have filters with different sizes $1 \times 1$, $3 \times 3$, $5 \times 5$ and $7 \times 7$, respectively as shown in Figure 2. These distinct sized filters enrich the capability of the network by extracting both local and abstracted features of the facial expression, which are the core attributes to define disparities between the expressions. Moreover, elective layer selects only exigent features from the former layer responses, by using Eq. 5-7.

$$R_\in = \frac{1}{2}\left((\max(R_n) + (\min(R_n))\right) \quad (5)$$

Where, $R_\in$ and $n$, is the regularized response of the previous responses and a total number of former incoming responses.

$$D_n = |R_\in - R_n| \quad (6)$$

Where, $D_n$ calculated distance between the response features and regularized response features.

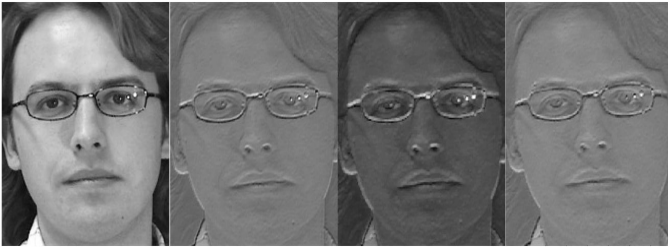

Figure 3: Input image and response images generated by applying (a) Conv with stride 1 (b) pooling with stride 2 and (c) Conv with stride 2.

$$R_E = R_\epsilon + \min(D_n) \qquad (7)$$

**Additive Layer:** Motivated from ResNet [16], EXPERTNet included additive layer's adaptability to improve the quality of responses. Additive layer accumulates the response feature map of each ExFeat block with outcome of previous convolution layer as shown in Figure 1. Additive layer enhances learnability of the hidden layer's neurons, which improves recognition accuracy of the network. The outcome of the additive layer is computed by using Eq. 8.

$$R_{add} = R_{Conv} + R_{ExFeat} \qquad (8)$$

Furthermore, EXPERTNet repeats the former described layer with distinct depth channels such as: 64, 96 and 128, to create the cross relationship between hidden layer's neurons. Cross relationship between layers helps to upgrade the learnable weights according to the facial appearance of expression. Then, updated feature maps are feeds to the fully connected layer.

**Fully Connected layer**: FC layer creates connection between every neuron of the former layers to each neuron of the self-generative layer. Thus, it is also known as multilayer perceptron. Let the input is z with the size of l and N represents a number of neurons presented at hidden layers. Then the activation function of the neurons is calculated by matrix multiplication added with bias. The activation function represents by Eq. 9.

$$f_z = \psi(M * z) \qquad (9)$$

Where, $\psi$ and $M_{N \times l}$, is the activation function and resultant matrix, respectively. The detailed configuration of the proposed architecture tabulated in Table. I.

### 2.1 Analysis of proposed network

Conventional CNN- based networks like AlexNet [13], VGG-Net [14], GoogleNet [15] and ResNet [16], gains attention in various fields eg: object detection, pattern analysis, face recognition, facial expression classification etc. Motivated by prior models, we proposed a novel architecture EXPERTNet to generates the feature maps of the facial expressions by adopting the linear cross correlation property along with dense depth of the applied filters. Dense deep network enhances the neurons learnability by updating the weights of corresponding filters. Moreover, EXPERTNet contains the ExFeat blocks, which

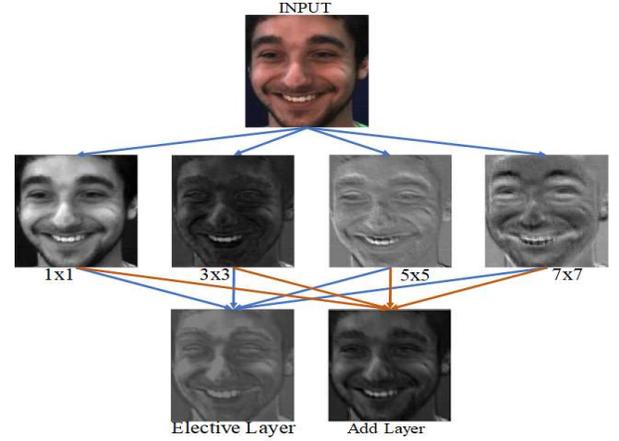

Figure 4: Input image and response feature maps generated by applying different sized filters as $1 \times 1$, $3 \times 3$, $5 \times 5$ and $7 \times 7$ in ExFeat block. Further, output feature map is generated by applying (a) elective layer and (b) addition layer, respectively. Visibly it is clear that elective layer preserves prominent edge features.

Incorporated different sized filters as $1 \times 1$, $3 \times 3$, $5 \times 5$ and $7 \times 7$. This combination of filters allows extracting both micro and high-level edge features. Further these features are feeds to the elective layer to preserve only exigent feature responses and processed to next layer, instead of all feature responses like Inception layer [15]. Feature extraction capability of elective layer is depicted in Figure 4, concerning addition layer. It improves the computation power and learning capability of the network. Further, resultant feature maps are combines by using additive layer, with the prior convolution layer to enrich the generated feature responses like ResNet [16]. These responses describe the discriminability among the various class facial expressions as shown in Figure 5. Figure 5, included the universal expressions as neutral, anger, disgust, fear, happy, sad and surprise of a single subject and their respective encoded responses. From the figure it is clear that, EXPERTNet successfully identified the differences between various expression classes

Moreover, EXPERTNet uses the convolution layer with stride 2, to reduce the size of input in-place of max pooling. Max pooling executed max function to scale down the input size, which ignores the minute edge features. But, in facial expression analysis, micro level edges are also played a significant role to describe facial appearance structure. Thus, EXPERTNet included convolution layer with stride 2, which decrease the size of input with minimum information loss as shown in Figure. 3.

## 3 Experimental Results and Analysis

The performance of proposed architecture has validated by using four benchmark datasets: CK+ [25], MMI [26, 27]. DISFA [28] and GEMEP-FERA [29]. These datasets are prepared to analyze the facial expressions under various challenges like posed and spontaneous expressions, ethnicity variations, a subject wearing different artifacts etc. We also evaluate performance of the proposed network for 7-class expression, where datasets included neutral expressions too. Furthermore, to extract the facial region, we have utilized the viola jones [30] face detection algorithm instead of manual face cropping like existing methods [11, 12]. This procedure creates a real-life scenario and enhances the adaptability of the model to deal with real-life conditions as images don't have static position and background. Moreover, to validate the outcomes, experimental setup included N-fold cross-validation scheme. In N- Fold cross-validation, image sets are partitioned into N equal sized folds, from which N-1 folds are included as training and remaining one is used as a test dataset. Further, to prepare datasets, we have extracted most expressive image frames and then divided them to 80:20 ratio as 80% for training and remaining for testing image set. To train the network, training set is again divided into 70:30 ratio, from which 70% images are used to train the network and 30% to validate the accuracy outcomes. The recognition rate for the network is calculated by using Eq 10.

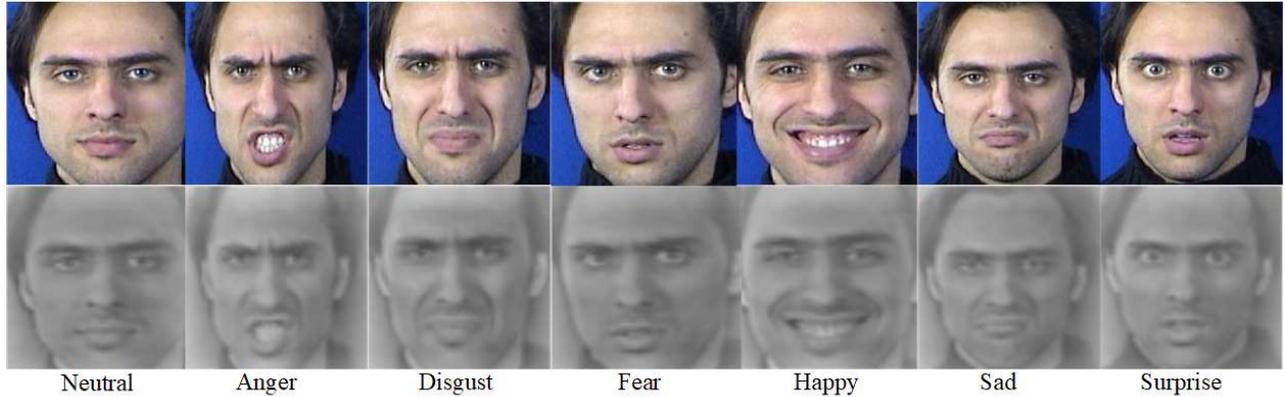

Figure 5: Visualization of response feature maps for a) neutral b) anger c) disgust d) fear e) happy f) sad and g) surprise expression, capture at elective layer over MMI dataset.

TABLE II
RECOGNITION ACCURACY COMPARISON ON CK+ DATASET

| Method | Accuracy Rate (%) | |
|---|---|---|
| | 6-class | 7-class |
| LBP [7] | 93.5 | 89.0 |
| Two-Phase [8] | 88.2 | 79.5 |
| LDP [9] | 96.2 | 92.9 |
| LDN [10] | 94.8 | 91.7 |
| LDTexP [11] | 95.3 | 91.9 |
| LDTerP [12] | 95.7 | 91.5 |
| VGG-Net 16 [14] | 96.7 | 95.2 |
| VGG-Net 19 [14] | 97.2 | 81.2 |
| ResNet [16] | 94.0 | 91.8 |
| **EXPERTNet** | **99.1** | **98.8** |

TABLE III
RECOGNITION ACCURACY COMPARISON ON MMI DATASET

| Method | Accuracy Rate (%) | |
|---|---|---|
| | 6-class | 7-class |
| LBP [7] | 76.5 | 81.7 |
| Two-Phase [8] | 75.4 | 82.0 |
| LDP [9] | 80.5 | 84.0 |
| LDN [10] | 80.5 | 83.0 |
| LDTexP [11] | 83.4 | 86.0 |
| LDTerP [12] | 80.6 | 80.0 |
| VGG-Net 16 [14] | 83.9 | 89.2 |
| VGG-Net 19 [14] | 81.6 | 83.9 |
| ResNet [16] | 71.2 | 83.9 |
| **EXPERTNet** | **99.1** | **98.0** |

$$Recog.\ Accur. = \frac{Total\ no.\ of\ correctly\ predicted\ samples}{Total\ no.\ of\ samples} \times 100 \quad (10)$$

To create a large dataset, we have performed augmentation by rotating each image in a random transaction range [-30, 30] along with x and y-axis to extend the small dataset into larger dataset. Moreover, to make fair comparison analysis between proposed and existing models, we also implemented existing methods according to our experimental setup The EXPERTNet have trained with 1e-3 learning rate, 35 min batch size and 200 epochs. Proposed algorithm is optimized using stochastic gradient descent (SGD). All networks are trained in Matlab 2018a with Titan-X GPU.

### 3.1 Experiments on CK+ Dataset

The CK [25] dataset is broadly used dataset for analysis of facial expression recognition systems. Initially, CK+ dataset comprises 489 image sequences of 97 posers. Further, CK dataset has been extended by including 593 image sequences for posed and non-posed expressions. The dataset contains cross ethnic subjects as American, African, Asian and Latin. In this dataset, each image sequence initiates with neutral and extend up to peak expression. All image sequences are labeled with one of six basic expressions: anger, disgust, fear, happy, sad and surprise. For the 7-class expression, we have included apex frame of each expression as neutral expression. In our experiment, we select 2897 total images with seven proper expression class labels as anger-343, disgust-419, fear-318, happy-577, neutral-326, sad-409 and surprise-505. Recognition accuracy results over ck dataset for existing state-of-art and EXPERTNet approach is detailed in Table II. By Table II, we can observe that, EXPERTNet gains more accuracy as compared to other existing FER approaches. Particularly, proposed network secure 2.4%, 1.9%, 5.1% and 3.6%, 17.6%, 7% more accuracy for 6-and 7-class expression as compare to VGG-Net 16, VGG-Net 19 and ResNet respectively. EXPERTNet also gain 3.8%, 3.4% and 6.9%, 7.3% more accuracy rate for 6-and 7-class expression over handcrafted approaches LDTexP and LDTerP respectively.

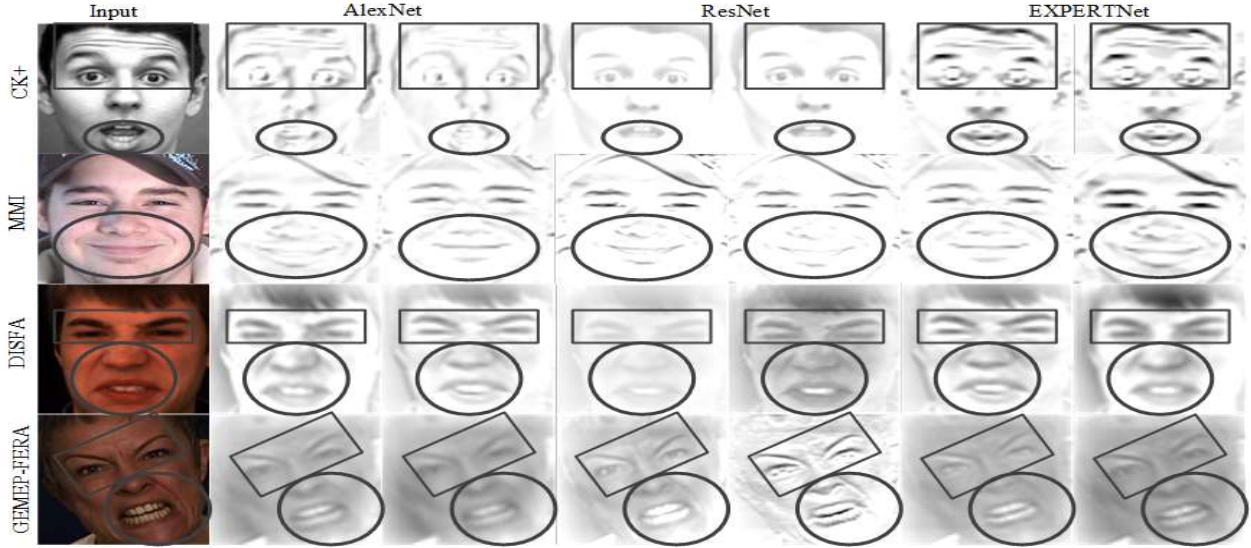

Figure 6: Visual comparison of existing models and EXPERTNet over different expression of four datasets a) CK+: Surprise b) MMI: Happy c) DISFA: Disgust and d) GEMEP-FERA: Anger.

TABLE IV
RECOGNITION ACCURACY COMPARISON ON DISFA DATASET

| Method | Accuracy Rate (%) | |
|---|---|---|
| | 6-class | 7-class |
| LBP [7] | 91.8 | 92.7 |
| Two-Phase [8] | 91.0 | 92.9 |
| LDP [9] | 91.5 | 94.1 |
| LDN [10] | 90.7 | 93.0 |
| LDTexP [11] | 92.2 | 93.8 |
| VGG-Net 16 [14] | 89.2 | 83.9 |
| VGG-Net 19 [14] | 83.9 | 88.3 |
| ResNet [16] | 83.9 | 71.2 |
| **EXPERTNet** | **95.3** | **95.5** |

### 3.2 Experiments on MMI Dataset

The MMI [26, 27] dataset included more than 2900 videos and frontal posed image sequences of 75 subjects. Each image sequence starts with neutral and go up to extensive expression, then again release with the neutral expression. In this dataset, subjects are captured with or without artifacts. The dataset contains both male and female subjects belonging to cross cultured environments as Europe Asia and South America. Similar to CK+, For MMI dataset, apex frame is considered as neutral expression to make a 7- class expression category. Our experimental setup included, total 3598 images with seven expression classes as: anger-483, disgust-501, fear-515, happy-615, neutral-456, sad-456 and surprise-572. Table III represents the comparative analysis of proposed network with existing methods. The effectiveness of proposed network is tabulated in Table. III in terms of recognition rate. Table. III evident that, EXPERTNet outperforms the existing FER approaches with sufficient margins. Specifically, EXPERTNet yields 15.2%, 17.5%, 27.9% and 8.8%, 14.1%, 14.1% accuracy rates as compare to CNN based networks: VGG-Net 16, VGG-Net 19 and ResNet for 6- and 7-class, respectively. Moreover, EXPERTNet also achieved 15.7%, 18.7% and 12.0%, 18.0% more accuracy as compared to conventional handcrafted descriptors: LDTexP and LDTerP for 6- and 7-class expressions, respectively.

### 3.3 Experiments on DISFA Dataset

The DISFA [28] dataset has over 8900 stereo video recordings of 27 subjects. The dataset included 12 male and 15 female subjects belonging to various origins. Mainly, this dataset created to analyze the spontaneous expression of the humans, which provide a real-time scenario to the FER application. Thus, subjects were captured while they were not aware of camera and watching some fascinating video clips. Finally, images are labeled with the six expression categories as: anger, disgust, fear, happy, sad and surprise, according 66 FACs. For experiment, we have elicited 2761 frames from the videos which have accurate expression labels as: anger-112, disgust-287, fear- 515, happy-674, neutral-615, sad-411 and surprise-265. Accuracy results over DISFA dataset are tabulated in Table IV. Table IV, shows that, proposed network outperformed the existing state-of-art approaches. Particularly, the proposed model gains 6.1%, 11.4%, 11.4% and 11.6%. 7.2%, 24.3% extra recognition accuracy for 6- and 7- class expression as compared to CNN based approaches VGG- Net 16, VGG- Net 19 and ResNet models respectively. Further, it yields 4.6%, 3.1% and 2.5%, 1.7% for 6-and 7-class expressions, better accuracy as compared to the handcrafted techniques of LDN and LDTexP respectively.

TABLE V
RECOGNITION ACCURACY COMPARISON ON GEMEP-FERA DATASET

| Method | Accuracy Rate (%) | |
|---|---|---|
| | 5-class | 6-class |
| LBP [7] | 92.2 | 87.8 |
| Two-Phase [8] | 88.6 | 85.0 |
| LDP [9] | 94.0 | 90.0 |
| LDN [10] | 93.4 | 91.0 |
| LDTexP [11] | 94.0 | 91.8 |
| VGG-Net 16 [13] | 85.1 | 90.7 |
| VGG-Net 19 [14] | 91.8 | 89.3 |
| ResNet [16] | 78.4 | 78.7 |
| **EXPERTNet** | **94.4** | **92.9** |

### 3.4 Experiments on GEMEP-FERA Dataset

The GEMEP-FERA [29] contains 226 videos of 10 subjects. The original dataset divided the images into two sets as 155 videos of 7 subjects comes under training set and 71 videos of 6 subjects are included in testing test. The dataset holds multiple sessions for a particular subject. In our experimental setup, we merged both datasets to maintain analogy in it with other experiments. Thus, final arranged dataset contains 725 images with six expression classes as: 142- anger, 132- fear, 141- joy, 77- neutral, 115- relief and 118- sad. Table V illustrates the effectiveness of the proposed EXPERTNet and other existing methods in terms of recognition accuracy. More specifically, our approach yields 67.0%, 0.1%, 1.9% and 73.4%. 2.9%, 6.7% better accuracy rate as for 6- and 7-class expression compared to VGG-Net 16, VGG-Net 19 and ResNet respectively. Moreover, it outperformed handcrafted methods: LDN and LDTexP by 1%, 0.4% and 1.9%, 1.1% for 6-and 7-class expressions respectively.

### 3.5 Qualitative Analysis

Figure 6 depicts the qualitative visual analysis of the existing and proposed model. This figure contains the visual representation of different emotion classes as *surprise, happy, disgust and anger* from all four datasets: CK+, MMI, DISFA and GEMEP-FERA respectively. In Figure 6, we have depicted two most prominent visual responses generated by intermediate hidden layers. Expressive regions like eyes, nose and mouth which play a significant role in defining disparities between emotion class are highlighted with black boxes. It is clear from Figure 6 that the response feature maps significantly assist in preserving the minute variations in different expressive regions in the facial image. For example, *in surprise: forehead*, mouth; *in happiness: eyes, mouth; in disgust: eyes, mouth and in anger: eyes, eyebrows mouth* regions give maximum affective response for related facial expressions. From Figure 6, we can conclude that EXPERTNet has preserved more relevant feature responses to outperform the existing CNN based networks AlexNet and ResNet for almost all emotion classes.

### 3.6 Computational Complexity

This section provides the comparative analysis of the computational complexity between the existing and proposed network. The proposed EXPERTNet has only 4M learnable parameters which are very less as compare to other existing benchmark models like: VGG-16: 138M, VGG-19: 144M, GoogleNet: 4M and ResNet: 11M. Moreover, EXPERTNet architecture has fewer depth channels and hidden layers as compared to former methods. Particularly, EXPERTNet comprises 13 layers. In comparison to that, VGG-16, VGG-19, GoogleNet and ResNet consists of 16, 19, 22 and 34 layers respectively.

## 4 Conclusion

This paper presents a new CNN architecture named as EXPERTNet: Exigent Features Preservative Network for facial expression recognition. EXPERTNet follows a linear cross-correlation behavior with deep dense convolution layers, which enhances its robustness to noise and ethnicity variations. EXPERTNet is designed in such a way that it has an ability to forward only prominent features to next layers by using ExFeat block. ExFeat blocks hold an elective layer, which elected the salient features and ignores remaining. Thus, it reduced the computational cost and enhanced performance of the EXPERTNet. Moreover, ExFeat block contains different sized filters as $1\times1$, $3\times3$, $5\times5$ and $7\times7$ to capture both local and abstracted features. Thereby, the response feature maps can easily extract the edge variations of facial appearance. Furthermore, EXPERTNet combines the former layer response with currently processed layer responses to secure more feature information. Thus, resultant feature maps have capability to define disparities between different expression classes. Experimental results have proved effectiveness of the proposed network over four datasets: CK+, MMI, DISFA and GEMEP-FERA.

**ACKNOWLEDGMENTS**

This work was supported by the Science and Engineering Research Board (under the Department of Science and Technology, Govt. of India) project #SERB/F/9507/2017. The author would like to thank our Vision Intelligence lab group for their valuable support. We are also thankful to NVIDIA for providing TITAN XP GPU grant.